\documentclass{article}
\usepackage{arxiv}

\usepackage[utf8]{inputenc} 
\usepackage[T1]{fontenc}    

\usepackage{graphicx}
\usepackage{amsmath} 
\usepackage{blkarray} 
\usepackage{multirow} 
\usepackage{amsfonts}
\usepackage{amssymb}
\usepackage[british]{babel} 
\usepackage{natbib} 
\usepackage[font={small}]{caption} 
\usepackage{subcaption} 
\usepackage{enumerate} 
\usepackage{chngcntr} 
\usepackage{hyperref}
\usepackage{url}            
\usepackage{nicefrac}       
\usepackage{microtype}      
\usepackage{lipsum}		
\usepackage{doi}
\usepackage[font=small,skip=2pt]{caption}
\usepackage{booktabs} 
\usepackage{tcolorbox}
\usepackage{cleveref} 

\DeclareMathAlphabet\mathbfcal{OMS}{cmsy}{b}{n} 

\author{ \href{https://orcid.org/0000-0002-3880-5440}{\includegraphics[scale=0.06]{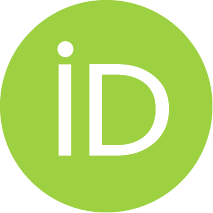}\hspace{1mm}Rogério Duarte}\thanks{Corresponding author.} \\
	CINEA, DEM, Escola Superior de Tecnologia de Setúbal\\
	Instituto Politécnico de Setúbal\\
	Campus do IPS, Estefanilha, 2914-761 Setúbal, Portugal \\
	\texttt{rogerio.duarte@estsetubal.ips.pt} \\
	\And
{Ângela Lacerda-Nobre} \\
	DEG, Escola Superior de Ciências Empresariais\\
	Instituto Politécnico de Setúbal\\
	Campus do IPS, Estefanilha, 2914-503 Setúbal, Portugal \\
	\texttt{angela.nobre@esce.ips.pt} \\
	\And
{Fernando Pimentel} \\
	DEM, Escola Superior de Tecnologia de Setúbal\\
	Instituto Politécnico de Setúbal\\
	Campus do IPS, Estefanilha, 2914-761 Setúbal, Portugal \\
	\texttt{fernando.pimentel@estsetubal.ips.pt} \\
	\And
{Marc Jacquinet} \\
	DCSG, Universidade Aberta\\
	Rua da Escola Politécnica, 141-147, 1269-001 Lisboa, Portugal \\
	\texttt{marc.jacquinet@uab.pt.} \\
}
\date{}

\renewcommand{\undertitle}{\phantom{undertitle}\par\phantom{undertitle}}
\renewcommand{\shorttitle}{Broader Terms Curriculum Mapping}

\hypersetup{
pdftitle={Broader Terms Curriculum Mapping: Using natural language processing and visual-supported communication to create representative program planning experiences},
pdfsubject={Using classification of learning objectives, natural language processing, and data visualization, this paper presents a method to deliver program plan representations that are universal, self-explanatory, and empowering},
pdfauthor={Rogério Duarte, Ângela Lacerda-Nobre, Fernando Pimentel, Marc Jacquinet},
pdfkeywords={Curriculum mapping, Natural Language Processing (NLP), Network graphs, Learning objectives classification, Curriculum analytics, Curriculum design, Higher education},
}

\title{Broader Terms Curriculum Mapping: Using natural language processing and visual-supported communication to create representative program planning experiences}
\begin{document}
\maketitle
\begin{abstract}
\noindent Accreditation bodies call for curriculum development processes open to all stakeholders, reflecting viewpoints of students, industry, university faculty and society. However, communication difficulties between faculty and non-faculty groups leave unexplored an immense collaboration potential. Using classification of learning objectives, natural language processing, and data visualization, this paper presents a method to deliver program plan representations that are universal, self-explanatory, and empowering. A simple example shows how the method contributes to representative program planning experiences and a case study is used to confirm the method's accuracy and utility.

\end{abstract}

\section{Introduction}\label{sec:1}
Changing times impose new and radical challenges for societies, urging Higher Education Institutions (HEI) to rethink their educational offer. Curriculum development is the process where changes to educational offer are conceived and, to be successful, this process needs to create opportunities for active and consequent reflection. To create these opportunities, stakeholders' participation is essential. This is the unanimous opinion of researchers and accreditation bodies \citep{webpageABET2020,Crawley2007,Sutherland2018} who defend curriculum development open to all stakeholder groups, expressing the viewpoints of students, industry, university faculty and society.

Focusing on program planning, a core process that lays at the heart of curriculum development, open principles are typically associated with interviews, focus group sessions, where non-faculty stakeholders are asked to express their views. Faculty, who hold the central managing role \citep{Wilson2020}, is responsible for processing the data collected in these informational touchpoints, and it is faculty who participate in program planning discussions.

Faculty central role in program planning is not only a traditional functional attribute. Program planning requires scientific and pedagogic skills essential to---and therefore mastered by---faculty, that are not essential to non-faculty stakeholders. This opens an important communication gap, and the absence of dialogue between faculty and non-faculty groups \citep{Ornstein2018} leaves unexplored an immense collaboration potential.

Yet, the challenges imposed by a rapidly changing society; recognizing that responsive and effective program plans are more likely with participatory (not just informational) involvement of all those concerned, forces HEI to reach out and find ways to integrate external contributions, valuing non-faculty stakeholders as experts of their own experience.

To achieve this objective, to bridge communication gaps, better representations of the program plan are essential. This paper proposes the use of a \textit{broader terms} curriculum mapping method to deliver program plan representations that are universal, self-explanatory, empowering.

The core objective of this paper is the presentation of the broader terms CM (Curriculum Mapping) method. Because this method uses a combination of information and data science techniques, a significant part of the paper is dedicated to the step by step description---using a simple example---of these techniques, and how they contribute to representative program planning experiences. To verify the method's accuracy and utility, a case study is used.

But, before proceeding to the broader terms CM method presentation, it is important to explain what is meant  by curriculum mapping. How it has been used previously, and what changes are required to make it a vehicle for representative program planning. This is the topic of the next section.

\section{The context: Curriculum mapping}\label{sec:2}
According to \citet{Burns2001}, curriculum mapping is a process for recording what content and skills are taught in a study program. The recording relies on a visual medium, typically a chart, table, or map, depicting the building blocks of the study program and how these blocks relate to one another. Because different types of building blocks could be used, there are different types of curriculum mapping.

When individual courses are the building blocks, curriculum mapping provides a snapshot of existing learning pathways considering the available courses, helping students navigate the study program. These \textit{course} mappings use the calendar year as an organizer to depict vertical (from year to year) and horizontal (within a year) relations between courses \citep{Burns2001}, and are usually represented as flowcharts. \citet{Meij2018} provide an example of a course mapping published online, with course-specific scientific and pedagogic details available as hyperlinked content.

For accreditation bodies, the grouping of contents and skills per courses is not as relevant as ensuring these contents and skills result in expected learning outcomes \citep{Felder2003,Crawley2007}. For this reason, for accreditation purposes, learning outcomes are the building blocks, and \textit{learning outcomes} mappings  are used to show the study program yields the expected learning outcomes. These mappings are typically represented as tables aligning program learning outcomes and accreditation standards. Examples of learning outcomes mappings are given in \citet{Dyjur2016}.

\textit{Learning outcomes} mappings' purpose goes beyond reporting the alignment with accreditation standards. This type of curriculum mapping is used to communicate accreditation bodies vision of transparency, accountability and scientific curriculum development \citep{Ornstein2018}. A vision that becomes reality with HEI adoption of outcomes-based education \citep{Spady1988,Harden2001} and constructive alignment principles \citep[p.99]{Biggs2011}. Curriculum mapping is used, therefore, as a tool to shape HEI processes, particularly, program planning.

\citeauthor{Willcox2017} describe another type of curriculum mapping: the \textit{concept} mapping. Concepts are in this case used for building blocks, with a concept denoting ``the main idea underlying a (typically small) unit of content covered in a course'' \citep[p.9]{Willcox2017}. These units of content are linked to \textit{Knowledge Concepts} defined by \citep{Koedinger2012}, and \citeauthor{Willcox2017} use concept mappings to provide insight into the relations between learning outcomes and between courses, helping faculty with the precise program plan navigation. Examples of this type of mapping are given in \citet{Seering2015}, \citet{Willcox2017} or \citet{Varagnolo2020}. These authors use circular ideograms \citep{Krzywinski2009} and/ or network graphs \citep{Rosen2009} to detail concepts' precedence relations. The visual outputs presented by these researchers are very successful and efficient in conveying visual meaning to the complex relations found study programs.

The analysis of the three types of curriculum mapping reveals important characteristics. Curriculum mapping is used to shape HEI processes, and this ability is valuable for opening program planning discussions to non-faculty groups. Curriculum mapping uses visual-supported communication to represent and discuss study programs, and the developments taking place in the field of information visualization can be used to bridge communication gaps, helping stakeholders to articulate their expert (non-verbal) knowledge. However, as regards the choice of building blocks, if the objective is to increase non-faculty groups participatory involvement, broader (not detailed) concepts, requiring less scientific and pedagogic skills should be preferred.

A curriculum mapping method that builds on the practices already available but tailored for non-faculty groups participation in program planning discussions is described in the next section.

\section{The method: Broader terms curriculum mapping}\label{sec:3}
This section presents a curriculum mapping method designed for representative program planning. A method that empowers all stakeholders.

A flowchart representing the method steps, respective inputs and outputs, is presented in Figure \ref{fig:1.0}.

\vspace{1\baselineskip}
\noindent%
\begin{minipage}{\linewidth}
  \center
  \includegraphics[scale=0.66]{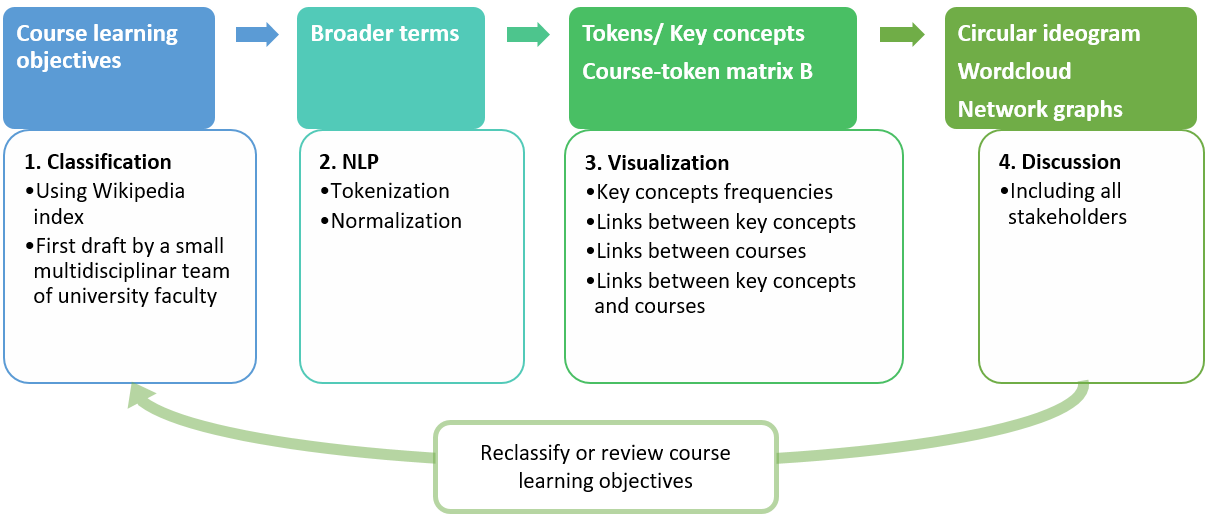}
  \captionof{figure}{Flowchart representing the steps, respective inputs and outputs of the broader terms curriculum mapping method.}\label{fig:1.0}
\end{minipage}

\vspace{1\baselineskip}
The method considers four steps---detailed in the following subsections---: (1) classification of course learning objective  statements into broader terms; (2) use of Natural Language Processing (NLP) to convert broader terms into quantitative frequencies of key program concepts; (3) visualization of key program concept frequencies and mappings with links between key concepts and/ or courses; (4) discussion, considering the participation of all stakeholder groups, of the method's visual outputs and decision to reclassify or review course learning objectives.

To illustrate how these steps apply, a simple example is considered. Table \ref{tab:1.0} presents data for this example.

\vspace{1\baselineskip}
\noindent%
\begin{minipage}{\linewidth}
  \center
  \captionof{table}{Learning objectives and respective broader terms classification for 5 courses. Notes: (1) A non-truncated version of this table is provided with the supplementary material \citep{Duarte2020}. (2) Data in this table models style and scope variability frequently found in learning objective statements and considers different levels of detail in broader terms selection.}\label{tab:1.0}
  \begin{tabular}{p{2cm}p{5.5cm}p{5.5cm}}
  \toprule
  \textbf{Course} & \textbf{Learning objectives} & \textbf{Broader terms}$^\mathbf{(a)}$ \\
  \midrule
  {\scriptsize Mathematics (C1:MATH)} & {\scriptsize Recognize a real-valued function of a real variable; (\ldots) Recall the concept of derivative of a real function and explain its geometric interpretation; (\ldots)} & {\scriptsize Function of a real variable; Differential \underline{calculus}; Integral \underline{calculus}; \underline{Linear} algebra; System of \underline{linear} equations} \\
  
  {\scriptsize Applied Physics (C2:PHY)} & {\scriptsize List fundamental concepts in mechanics and understand their importance to engineering; Use the international units (\dots)} & {\scriptsize Physics; Mathematics; \underline{Calculus}; Mechanics; Thermodynamics; Fluid flow} \\

  {\scriptsize Logistics~\&~Oper. Manag. (C3:LOGOP)} & {\scriptsize Identify logistic activities in a generic organization;  Explain the role of contemporary logistics; (\ldots) Distinguish components of supply chain management (\ldots)} & {\scriptsize Logistics; Supply chain management; Business; Production Economics; Operations management; \underline{Linear} programming; Lean manufacturing; Process Control (\ldots)} \\

  {\scriptsize Energy Manag. (C4:ENER)} & {\scriptsize Discuss the efficient use of energy in industry, buildings and transports; Recognize applicable legislation and defend energy efficiency as (\ldots)} & {\scriptsize Energy efficiency; Organization; Buildings; Facility management; Logistics; Production planning and control; Solar water heating (\ldots).} \\

  {\scriptsize Financial Manag. (C5:FIN)} & {\scriptsize List fundamental financial management concepts and functions; Recognize and explain financial statements;  Contrast the economic and the financial analysis (\ldots)} & {\scriptsize Financial management; Accounting; Economics; Finance; Organization; Business governance; Corporate law; Trade; Return on invested capital (\ldots)} \\

  \bottomrule

  \multicolumn{3}{l}{\scriptsize (a) Obtained with the Wikipedia index \citep{wiki:indices}.} \\
  \end{tabular}

\end{minipage}

\vspace{1\baselineskip}
Table \ref{tab:1.0} includes the learning objectives for 5 courses---Mathematics (C1:MATH), Applied Physics (C2:PHY), Logistics~\&~Operations Management (C3:LOGOP), Energy Management (C4:ENER) and Financial Management (C5:FIN)---of a bachelor degree in Technology and Industrial Management (also used in the case study section). For example, the first learning objective statement in the Mathematics (C1:MATH) course is: ``Recognize a real-valued function of a real variable''.

The third column of Table \ref{tab:1.0} presents broader terms derived from the courses learning objectives. The methodology used to obtain these broader terms is described in the following subsection.

\subsection{Step 1: Classification of course learning objectives}\label{sec:3.1}
To characterize courses and the program-degree, the broader terms CM method uses course Learning Objectives (LO). According to \citet[p.19]{Felder2003}, course LO are defined as ``statements of observable actions that serve as evidence of the knowledge, skills and attitudes acquired in a course.'' These statements define key program concepts and through these key concepts the intricate web of course relations is revealed. Course LO provide, therefore, access to the ``mechanics'' behind a program plan.

The problem of using course LO is that they presume tacit understanding of concepts specific to disciplinary and scientific sub-areas, and this renders LO-statements seldom clear and unequivocal \citep{Ballantyne2019,Watts2015}. Even when LO are written according to specific rules (e.g., considering Bloom's taxonomy, \citealp{Bloom1956,Adam2004,Felder2003}), the variability in style and scope results in a heterogeneous set, including statements that are often too abstract or too detailed \citep{Lam2016,Hussey2003}.

To disclose their latent information and for effective communication, LO-statements would benefit from techniques used by library and information science professionals in resource classification. Resource classification indicates what a resource is about, and to achieve this goal a control vocabulary, a set of broader terms (concepts or subject headings, \citealp{Lancaster2003}) supporting classification has to be agreed upon. Control vocabularies are usually chosen among bibliographic classification schemes (such as the Dewey Decimal Classification), lists of subject headings \citep{LibraryCongressS/D} and thesauri \citep{EUROVOCS/D,UNESCOS/D,IEEE2019}. More recently, for its comprehensiveness and up-to-dateness, the Wikipedia index \citep{wiki:indices} is also used (see \citealp{Joorabchi2013} and \citealp{Bergman2015} for a discussion of the advantages of using Wikipedia's index as control vocabulary).

This paper considers principles of resource classification to classify course LO. Concepts from Wikipedia index matching course LO-statements are used to define broader terms. To illustrate how this is done, consider the excerpt of LO statements for Mathematics (C1:MATH) in Table \ref{tab:1.0}: ``Recognize a real-valued function of a real variable; Recall the concept of derivative of a real function and explain its geometric interpretation.'' Using Wikipedia index, the first statement could be classified according to the Wikipedia concept, ``Function of a real variable''; a matching concept almost identical to the original LO. However, broader concepts could be chosen. For example, the second LO statement could be classified (with exaggeration) by the broader Wikipedia concept ``Differential calculus''. The third column in Table \ref{tab:1.0} includes results of course LO-statements classification for the 5 courses.

The classification example provided in the previous paragraph, and the comparison of columns two and three in Table \ref{tab:1.0}, shows a significant reduction in the vocabulary used to characterize the courses. Because the reduction in vocabulary could entail an important loss of information, it follows the importance of the conceptual analysis of LO-statements \citep{Lancaster2003}. The importance of determining what the LO statement is about---the ``aboutness''---and of the translation into (the selection of) specific broader terms.

Because program planning relies, typically, on the declared curriculum, with LO-statements written by university faculty, the classification of LO is frequently done by faculty in collaboration with curriculum planners \citep{Willcox2017,Seering2015,Varagnolo2020}. An alternative procedure consists of an initial classification by an information science professional, subsequently revised by faculty \citep{Ballantyne2019}. For large database classification, automated machine learning techniques are also used \citep{GolubS/D,West2016}.  In this paper, an initial draft classification of course LO is made by a small multidisciplinary team of university faculty. Once visual outputs derived with the broader terms CM method are available, a reclassification is made with contributions from all stakeholders (see Step 4 in Figure \ref{fig:1.0}).

For effective communication of the program plan, having courses associated with a small subset of broader terms selected from a control vocabulary is an important advantage. Key concepts found in courses can be identified, paving the way to their quantification and to the analysis of the relations between courses, i.e., to the analysis of information flows, such as topics covered, which assessments relate to which topic, and so on.

The next section describes in detail the method used in the quantitative processing of broader terms.

\subsection{Step 2: Processing of broader terms}\label{sec:3.2}
This paper uses Natural Language Processing (NLP, \citealp{Manning1999,West2018}) to convert broader terms assigned to courses into quantitative data; i.e., into frequencies of words. It will be assumed that these words---these tokens as they are called in the NPL literature \citep{Manning1999}---extracted from broader terms, still carry conceptual meaning and can still be used to characterize courses and the program-degree. For this reason, in this paper, \textit{token} and \textit{key} (program or course) \textit{concept}, K, are used as synonyms.

NLP applies a sequence of processing functions to an original set of broader terms. Tokenization, the first of these functions, identifies words in broader terms that are included in a corpus; in a dictionary of tokens. Recalling Section \ref{sec:3.1} example of obtaining broader terms for Mathematics (C1:MATH)---``Function of a real variable'' and ``differential calculus'' were the resulting broader terms---, and, considering the corpus of English words; after tokenization the following set of tokens \(~\lbrace \text{Function, of, a, real, variable, differential, calculus}\rbrace\) characterizes the Mathematics course.

But the above set includes tokens (i.e., ``of'' and ``a'') that add no value to the course characterization; therefore, these tokens---known as stop-words---, as well as any punctuation signs and numerals, should be removed. Moreover, words written with capital letters and different conjugations of the same word should be replaced by an adequate ``stem-word'' (in a process known as stemming, \citealp{Manning1999}).

Denoting the stemming and the purging of meaningless tokens as normalization, if a study program has \(N\) courses, after tokenization and normalization of course $\text{C}i$ broader terms ($i\in\lbrace 1,2,\ldots, N\rbrace$), a multiset $\mathbfcal{K}_i$ (allowing multiple instances of the same token) of $m_i$ tokens $\mathcal{K}_{i,k}$ is obtained (with $k=1,2,\ldots, m_i$). For the program-degree as a whole, a set $\mathbf{K}$ (no repetitions) with a total of $M=\lvert\bigcup_{i=1}^{N}\mathbfcal{K}_i \rvert$ tokens is obtained.

With $\text{K}j$ the $j^\text{th}$ token in set $\mathbf{K}$, the frequency of this token in course $\text{C}i$ is found from,
\begin{equation}\label{Eq:1}
  b_{i,j}=\sum_{k=1}^{m_i}\delta_{i,k}(\text{K}j)~,
\end{equation}

\noindent with 
\begin{equation}\label{Eq:2}
  \delta_{i,k}(\text{K}_j)=\left\{\begin{matrix}
0, & \mathcal{K}_{i,k}\ne \text{K}j \\
1, & \mathcal{K}_{i,k}= \text{K}j
\end{matrix}\right.~.
\end{equation}

Eq.~(\ref{Eq:1}) represents the elements of an $N\times M$ matrix $\mathbf{B}$ of token frequencies per course.

Considering the broader terms for the 5 courses in Table \ref{tab:1.0} (column three), after tokenization and normalization\footnote{These functions are available in most mathematical and numerical computing tools. \texttt{R} programming code \citep{R2019} used in this section is included in supplementary material, \citep{Duarte2020}.}, the resulting course-token matrix is,

\vspace{1\baselineskip}
\begin{footnotesize}
\begin{equation}\label{Eq:3}
  \mathbf{B_\text{5C}}=
\begin{blockarray}{cccccccc}
\text{\scriptsize{K1:manag}} & \text{\scriptsize{K2:calculus}} & 
\text{\scriptsize{K3:control}} & \text{\scriptsize{K4:energi}} & \text{\scriptsize{K5:linear}} & \text{\scriptsize{K6:logistic}} & \text{\scriptsize{(\ldots)}} & \\
\begin{block}{(ccccccc)c}
  0 & 2 & 0 & 0 & 2 & 0 & (\ldots) & \text{\scriptsize{C1:MATH}} \\
  0 & 1 & 0 & 0 & 0 & 0 & (\ldots) & \text{\scriptsize{C2:PHY}} \\
  2 & 0 & 2 & 1 & 1 & 2 & (\ldots) & \text{\scriptsize{C3:LOGOP}} \\
  3 & 0 & 1 & 2 & 0 & 1 & (\ldots) & \text{\scriptsize{C4:ENER}} \\
  1 & 0 & 0 & 0 & 0 & 0 & (\ldots) & \text{\scriptsize{C5:FIN}} \\
\end{block}
\end{blockarray}~,
\end{equation}
\end{footnotesize}

\noindent where, given the large number of identified tokens (70), only the columns for the six most frequent are shown.

Observe how this matrix attaches quantitative information to courses based on token frequency. Observe, for instance, the link that emerges between courses C1:MATH and C2:PHY via token K2:calculus. Matrix $\mathbf{B_\text{5C}}$ shows this token is found twice among the tokens associated with course C1:MATH, and once among those associated with course C2:PHY (see also the underlined words in Table \ref{tab:1.0}). 

This ability to describe a study program quantitatively is an important breakthrough; a way to bridge the gap created by tacit understanding, unclear LO-statements. But, at the same time, notice how unpractical is the analysis of the data in matrix format.

To achieve a clearer understanding of the quantitative data emerging from NLP, an alternative to matrix or tabular representations of data is essential.

\subsection{Step 3: Visualization of quantitative data}\label{sec:3.3}
An important result from NLP are token frequencies: the column-wise sum of the elements in the course-token matrix. A convenient visual representation of these frequencies is obtained with wordclouds. Figure \ref{fig:2.0} presents a wordcloud from data in matrix  $\mathbf{B}_\text{5C}$ (Eq.~\ref{Eq:3}).

\vspace{1\baselineskip}
\noindent%
\begin{minipage}{\linewidth}
  \center
  \includegraphics[scale=0.50]{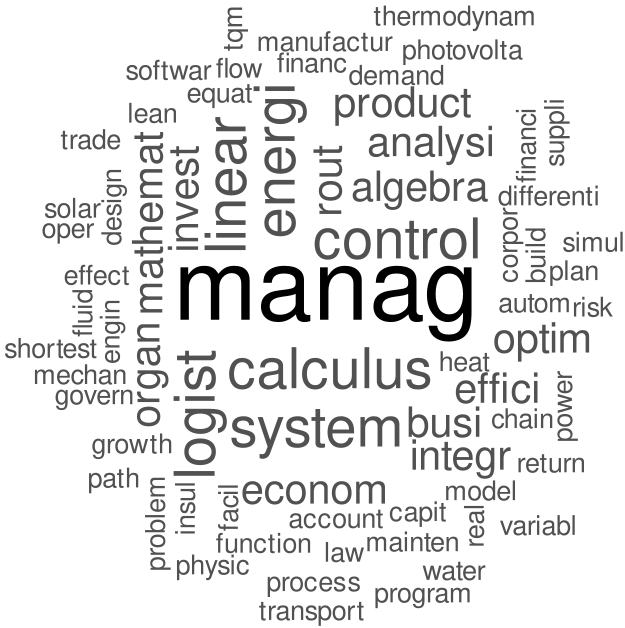}
  \captionof{figure}{Wordcloud of token frequencies for the 5 courses example. Graph obtained using the \texttt{Wordcloud} package \citep{wordcloud2018} for the \texttt{R} programming language \citep{R2019}---see supplementary material \citep{Duarte2020}.}\label{fig:2.0}
\end{minipage}

\vspace{1\baselineskip}
Figure \ref{fig:2.0} identifies the most frequent key program concepts---manag[ement], calculus, control, energi, linear, logistic---, represented with larger font size in a central position.

Figure \ref{fig:2.0} is adequate to identify the relative importance of different key concepts, but provides no information concerning the relations between these or between courses. To represent these relations researchers can choose among several alternatives. One which captures all data in course-token matrix and makes patterns and descriptive statistics visible is presented in Figure \ref{fig:3.0} a). It is the circular ideogram representation \citep{Krzywinski2009} of the data in matrix $\mathbf{B}_\text{5C6K}=\mathbf{B}_\text{5C}\left[1:5;1:6\right]$, a submatrix including the first 6 columns of matrix $\mathbf{B}_\text{5C}$ (Eq.~\ref{Eq:3}).

\vspace{1\baselineskip}
\noindent%
\begin{minipage}{\linewidth}
  \center
  \includegraphics[scale=1.25]{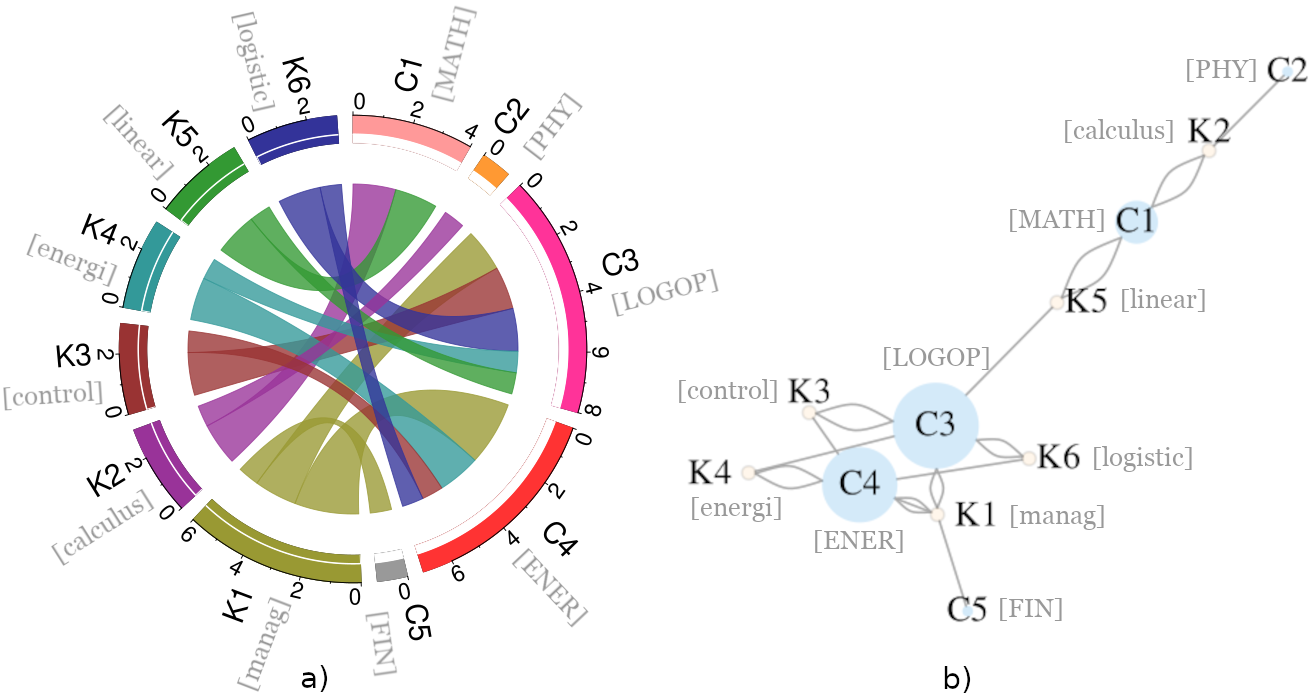}
  \captionof{figure}{Visual representation of matrix $\mathbf{B}_\text{5C6K}$ as: (a) a circular ideogram; (b) a multigraph. The circular ideogram \citep{Krzywinski2009} was obtained with the \texttt{Circlize} package \citep{Gu2014} and the multigraph was obtained with the \texttt{iGraph} package \citep{igraph2006}. Both packages for the \texttt{R} programming language \citep{R2019}---see supplementary material \citep{Duarte2020}.}\label{fig:3.0}
\end{minipage}

\vspace{1\baselineskip}
The outer circumference in Figure \ref{fig:3.0} a) displays the 5 courses $\text{C}i$ on the right side and the 6 tokens $\text{K}j$ on the left side. This circumference specifies the number of links (see scale) between courses and tokens. For example, courses C3 and C4 have the largest number of links (8 and 7, respectively) to tokens. Token K1 has the largest number of links (6) to courses.

But the advantage of the circular ideogram comes, especially, from the inner circle in Figure \ref{fig:3.0} a), and from the stripes that link courses and tokens. The inner circle in Figure \ref{fig:3.0} a) emphasizes the (previously mentioned) link between courses C1:MATH and C2:PHY via token K2:calculus (see purple stripe). But much more is revealed: for example, while course C2:PHY has no further associations, course C1:MATH is also related to course C3:LOGOP through K5:linear (green stripe). The width of the stripes---the strength of the links---connecting C1:MATH to K2:calculus and K5:linear is also larger than the width for the stripes connecting these tokens to courses C2:PHY and C3:LOGOP. Given that ``calculus'' and ``linear'' are mathematics-related tokens, these results were expected, and the expert analysis of the 5 courses LO-statements (in Table~\ref{tab:1.0}) should result in identical conclusions. But in Figure \ref{fig:3.0}, the combination of stripes' curvature, color and width renders the analysis universal, self-explanatory, empowering, uncovering latent information and helping the verbal articulation of expert (non-verbal) knowledge.

Another equally useful visual representation of data in matrix $\mathbf{B}_\text{5C6K}$ is presented in Figure \ref{fig:3.0} b). Consider the $\text{C}i$ and $\text{K}j$ in the outer circumference of Figure \ref{fig:3.0} a) as vertices $V=\lbrace v_\text{C1},v_\text{C2},\ldots,v_\text{K1},v_\text{K2},\ldots\rbrace$, and the inner circle stripes as edges $E=\lbrace e_{\text{C1-K2}},e_{\text{C1-K5}},\ldots\rbrace$ of an undirected multigraph $G=\langle V, E\rangle$. Figure \ref{fig:3.0} b) represents this multigraph with course and token vertices laid out in a way that communicates vertex centrality, i.e., where the number (the cardinality) of vertex links determine the vertex position \citep{Fruchterman1991}. Notice how vertices C3 and C4---with largest number of links---shape a central cluster, while vertices C1 and C2 protrude to the periphery. Moreover, vertex centrality is emphasized through course vertices diameter; with larger diameters representing courses with a larger number of incident links.

In matrix format, the multigraph in Figure \ref{fig:3.0} b) for the 5 courses and 6 most frequent tokens is,

\begin{footnotesize}
\begin{equation}\label{Eq:4}
\mathbf{A}_\text{5C6K}=
\left(\begin{array}{c|c}
   	\mathbf{0} & \mathbf{B}_\text{5C6K} \\
   	\midrule
   	\mathbf{B}_\text{5C6K}^T & \mathbf{0} \\
\end{array}\right)=
\begin{blockarray}{cccccccccccc}
\text{\scriptsize{C1}} & \text{\scriptsize{C2}} &
\text{\scriptsize{C3}} & \text{\scriptsize{C4}} & \text{\scriptsize{C5}} & \text{\scriptsize{K1}} & \text{\scriptsize{K2}} & \text{\scriptsize{K3}} & \text{\scriptsize{K4}} & \text{\scriptsize{K5}} & \text{\scriptsize{K6}} & \\
\begin{block}{(ccccc|cccccc)c}
\BAmulticolumn{5}{c|}{\multirow{5}{*}{$\mathbf{0}$}} & 0& 2& 0& 0& 2& 0& \text{\scriptsize{C1}} \\
&  &  &  &  &  0& 1& 0& 0& 0& 0& \text{\scriptsize{C2}} \\
&  &  &  &  &  2& 0& 2& 1& 1& 2& \text{\scriptsize{C3}} \\
&  &  &  &  &  3& 0& 1& 2& 0& 1& \text{\scriptsize{C4}} \\
&  &  &  &  &  1& 0& 0& 0& 0& 0& \text{\scriptsize{C5}} \\\cline{1-11}
0& 0& 2& 3& 1& \BAmulticolumn{6}{c}{\multirow{6}{*}{$\mathbf{0}$}} & \text{\scriptsize{K1}} \\
2& 1& 0& 0& 0& & & & & & & \text{\scriptsize{K2}} \\
0& 0& 2& 1& 0& & & & & & & \text{\scriptsize{K3}} \\
0& 0& 1& 2& 0& & & & & & & \text{\scriptsize{K4}} \\
2& 0& 1& 0& 0& & & & & & & \text{\scriptsize{K5}} \\
0& 0& 2& 1& 0& & & & & & & \text{\scriptsize{K6}} \\
\end{block}
\end{blockarray}~,
\end{equation}
\end{footnotesize}

\noindent a square biadjacency matrix obtained from $\mathbf{B}_\text{5C6K}$ (superscript $T$ denotes matrix transpose).

Figures \ref{fig:3.0} a) and \ref{fig:3.0} b) provide important insights on how key concepts and courses interrelate. However, a simpler and yet very useful representation would consist of the direct links between courses and between tokens.

Observing Figure \ref{fig:3.0} b) and matrix $\mathbf{A}_\text{5C6K}$ we conclude that elements of the biadjacency matrix represent the cardinality of 1-walks between consecutive vertices---with a $k$-walk defined as the sequence of $k$ edges $\left(e_1,e_2,\ldots, e_k\right)$ joining $k+1$ vertices $\left(v_1,v_2,\ldots, v_{k+1}\right)$ \citep{Rosen2009}. For example, matrix $\mathbf{A}_\text{5C6K}$ shows that between vertices $v_\text{C1}$ and $v_\text{K2}$ there are two 1-walks, $v_{\text{C1}}\xrightarrow{2\cdot e_{\text{C1-K2}}}v_{\text{K2}}$. Between vertices $v_\text{K2}$ and $v_\text{C2}$ there is one 1-walk, $v_{\text{K2}}\xrightarrow{1\cdot e_{\text{K2-C2}}}v_{\text{C2}}$. This is confirmed in Figure \ref{fig:3.0} b).

A direct link between vertices $v_\text{C1}$ and $v_\text{C2}$ could be conceived as two 2-walks joining these vertices, represented as $v_{\text{C1}}\xrightarrow{2\cdot e_{\text{C1-K2-C2}}}v_{\text{C2}}$, with $2\cdot e_{\text{C1-K2-C2}}$ denoting the two available options to go from C1 to C2.

For the 5 courses example, using matrix algebra, the number of 2-walks between course vertices and between token vertices is found from the 2\textsuperscript{nd} power of the biadjacency matrix, with the diagonal elements of the resulting matrix made equal to zero \citep{Rosen2009}. With $\mathbf{L}_\text{5C6K}$ denoting the 2-walk matrix, it follows $\mathbf{L}_\text{5C6K}=\mathbf{A}_\text{5C6K}^2-diag\left(\mathbf{A}_\text{5C6K}^2\right)$, and replacing $\mathbf{A}_\text{5C6K}$ gives,

\begin{footnotesize}
\begin{equation}\label{Eq:5}
  \mathbf{L}_\text{5C6K}=
\begin{blockarray}{cccccccccccc}
\text{\scriptsize{C1}} & \text{\scriptsize{C2}} &
\text{\scriptsize{C3}} & \text{\scriptsize{C4}} & \text{\scriptsize{C5}} & \text{\scriptsize{K1}} & \text{\scriptsize{K2}} & \text{\scriptsize{K3}} & \text{\scriptsize{K4}} & \text{\scriptsize{K5}} & \text{\scriptsize{K6}} & \\
\begin{block}{(ccccc|cccccc)c}
0& 2& 2& 0& 0& \BAmulticolumn{6}{c}{\multirow{5}{*}{$\mathbf{0}$}} & \text{\scriptsize{C1}} \\
2& 0& 0& 0& 0& & & & & & & \text{\scriptsize{C2}} \\
2& 0& 0&12& 2& & & & & & & \text{\scriptsize{C3}} \\
0& 0&12& 0& 3& & & & & & & \text{\scriptsize{C4}} \\
0& 0& 2& 3& 0& & & & & & & \text{\scriptsize{C5}} \\\cline{1-11}
\BAmulticolumn{5}{c|}{\multirow{6}{*}{$\mathbf{0}$}} & 0& 0& 7& 8& 2& 7& \text{\scriptsize{K1}} \\
&  &  &  &  &  0& 0& 0& 0& 4& 0& \text{\scriptsize{K2}} \\
&  &  &  &  &  7& 0& 0& 4& 2& 5& \text{\scriptsize{K3}} \\
&  &  &  &  &  8& 0& 4& 0& 1& 4& \text{\scriptsize{K4}} \\
&  &  &  &  &  2& 4& 2& 1& 0& 2& \text{\scriptsize{K5}} \\
&  &  &  &  &  7& 0& 5& 4& 2& 0& \text{\scriptsize{K6}} \\
\end{block}
\end{blockarray}
=\left(\begin{array}{c|c}
   	\mathbf{L}_\text{5C} & \mathbf{0} \\
   	\midrule
   	\mathbf{0} & \mathbf{L}_\text{6K} \\
\end{array}\right)~.
\end{equation}
\end{footnotesize}

Submatrices $\mathbf{L}_\text{5C}=\mathbf{L}_\text{5C6K} \left[1:5;1:5\right]$ and $\mathbf{L}_\text{6K}=\mathbf{L}_\text{5C6K} \left[6:11;6:11\right]$ in Eq.~(\ref{Eq:5}) represent the number of possible 2-walks between consecutive courses and consecutive tokens, respectively.

To confirm the results discussed previously for the direct link between vertices $v_\text{C1}$ and $v_\text{C2}$, notice the value 2 found in matrix element $\mathbf{L}_\text{5C6K} \left[1;2\right]$ (or $\mathbf{L}_\text{5C6K} \left[2;1\right]$, because the graph is undirected).

Using submatrices $\mathbf{L}_\text{5C}$ and $\mathbf{L}_\text{6K}$, representations of the direct links between courses and between key concepts are presented in Figures \ref{fig:4.0} a) and \ref{fig:4.0} b), respectively. The strength of the links---the cardinality of possible 2-walks---is given both by numbers and by edge widths. Moreover, as for Figure \ref{fig:3.0} b), vertices layout and vertex diameter provide a suggestive visual depiction of core and peripheral courses/ key concepts.

\vspace{1\baselineskip}
\noindent%
\begin{minipage}{\linewidth}
  \center
  \includegraphics[scale=1.330]{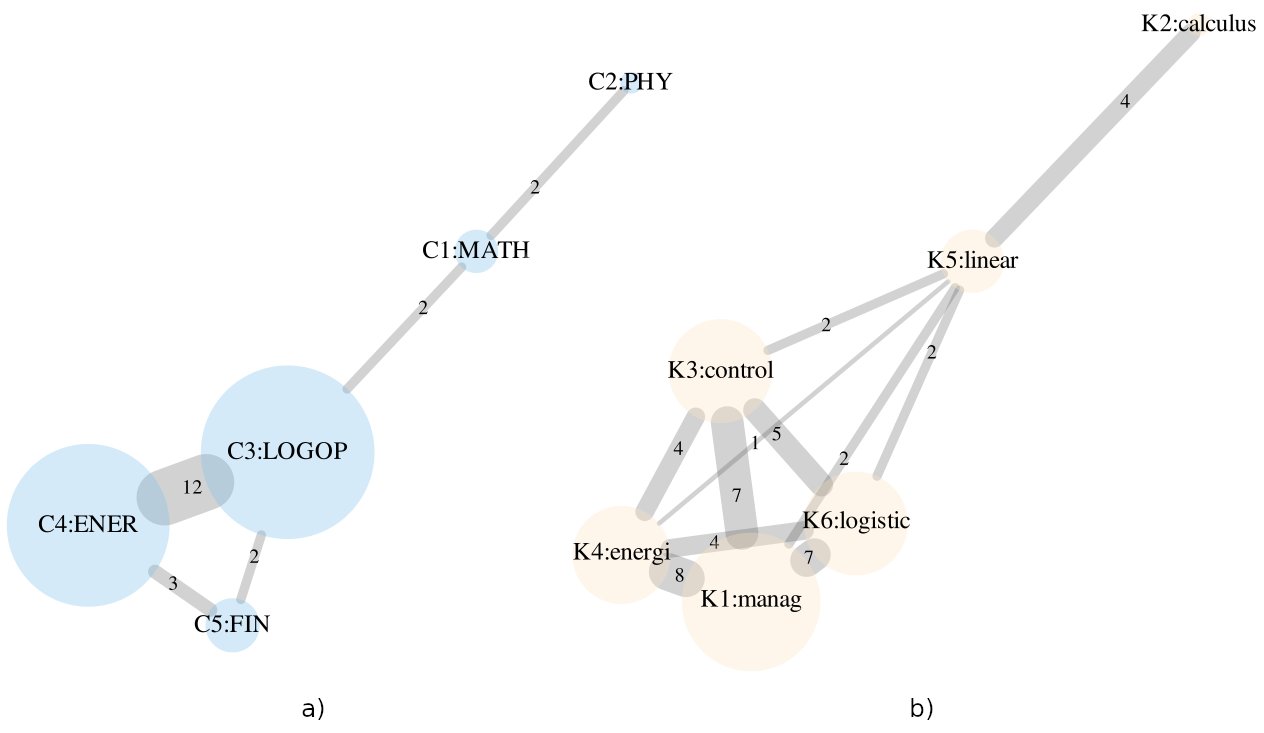}
  \captionof{figure}{Graphs showing direct links between: a) courses (from \(\mathbf{L}_\text{5C}\)); b) key concepts (from \(\mathbf{L}_\text{6K}\)). Numbers and edge widths represent the strength of the link. Graphs produced with the \texttt{iGraph} package \citep{igraph2006} for the \texttt{R} programming language \citep{R2019}---see supplementary material \citep{Duarte2020}.}\label{fig:4.0}
\end{minipage}

\vspace{1\baselineskip}
Figure \ref{fig:4.0} a) shows the largest number of possible 2-walks (12) occurs between courses C3:LOGOP and C4:ENER. This value can be verified in Figure \ref{fig:3.0} b). The way key concepts influence links between courses is clearly reflected in courses C2:PHY and C5:FIN locations. Although C2:PHY and C5:FIN have both a single key concept among the 6 most frequent---K2:calculus and K1:manag, respectively (see Eq.~\ref{Eq:3})---, the fact that ``manag[ment]'' is more common than the mathematics-related concept pulls C5:FIN closer to where core program courses lay, whereas C2:PHY is pushed to a peripheral location.

Figure \ref{fig:4.0} b) confirms the peripheral role played by mathematics-related concept, K2:calculus; and it is interesting to contrast this graph's discriminating potential with that of the wordcloud in Figure \ref{fig:2.0}. Indeed, no evidence is found in the wordcloud as to differences between tokens K2 to K6 (because the number of edges incident on vertices $v_\text{K2}$ to $v_\text{K6}$ is the same for these tokens, 3).

Figures \ref{fig:4.0} a) and Figure \ref{fig:3.0} b) provide visual evidence of course C2:PHY detachment from the remaining courses. Obviously, reasons for this should be discussed; in particular, the absence of a (expected) link between C2:PHY and C4:ENER.  

Results from this section show visual outputs from the broader terms CM method provide evidence-based details on weaknesses (and strengths) in program plans; namely, related to key program concepts and to the interrelations between these and/ or courses.

\subsection{Step 4: Discussion of the visual outputs}\label{sec:3.4}
With the adoption of the broader terms CM method the focus of program planning discussions is shifted from the discussion of written statements of course LO---seldom clear and unequivocal---; from atomized discourses about the links between courses, to the interpretation of \textit{quantitative} data communicated \textit{visually} in a way understandable to all.

Because of the universal, self-explanatory quality of its \textit{visual outputs}---of the mappings---, the broader terms CM method empowers all stakeholders, allowing participatory involvement of non-faculty groups in program planning discussions. Because of its \textit{quantitative} nature, the broader terms CM method nurtures constructive critique, effectively addressing disciplinary and scientific boundaries, hierarchical and functional differences, and atomized discourses. With the objective identification of program plan weaknesses, it is possible to unfreeze \citep{Schein1999} long established beliefs, preparing the agreement for change with contributions from all stakeholders (see the review feedback loop in Figure \ref{fig:1.0}).

Some of the weaknesses identified in the mappings may derive from course LO classification. Section \ref{sec:3.1} stated an initial draft classification was made by a small multidisciplinary team of university faculty. During the discussion step, with the help of mappings, classification problems are easily identified, justifying the reclassification feedback loop in Figure \ref{fig:1.0}.

As mentioned in Section \ref{sec:3.1}, this reclassification carries some subjectivity. Different broader terms could be chosen to classify a LO-statement; and there could be LO-statements for which an adequate broader term is not included in the control vocabulary. However, the technical nature of control vocabularies and of the classification task makes selection of broader terms distinctly less subjective than the head-on discussion of LO-statements.

Using the mappings for the 5 courses example we elaborate on relevant discussion topics that would benefit from the participatory involvement of all stakeholders. 

Considering frequencies and links between key program concepts, in Figures \ref{fig:3.0} a) and \ref{fig:4.0} b), stakeholders (namely, industry and society groups) could contribute with their experience to identify important key concepts, essential links, considering not only scientific and pedagogic arguments, but also the mission of HEI in the context of rapidly changing technological, economical, societal and political environments. With respect to the links between courses, and the links between courses and key concepts, in Figures \ref{fig:4.0} a) and \ref{fig:3.0} b), student and graduate groups could contribute with their experience to contrast the differences between the declared and the enacted curriculum \citep{Arafeh2016,Varagnolo2020}.

Concerning the lack of an expected link between C2:PHY and C4:ENER, mentioned at the end of last subsection---recall Figure \ref{fig:4.0} a)---; given that Applied Physics and  Energy Management syllabuses are typically linked by thermodynamics, heat transfer and fluid flow topics, to express the importance of this link, stakeholders could use handwritten notes to communicate a desirable change, as depicted in Figure \ref{fig:5.0}. 

\vspace{1\baselineskip}
\noindent%
\begin{minipage}{\linewidth}
  \center
  \includegraphics[scale=0.30]{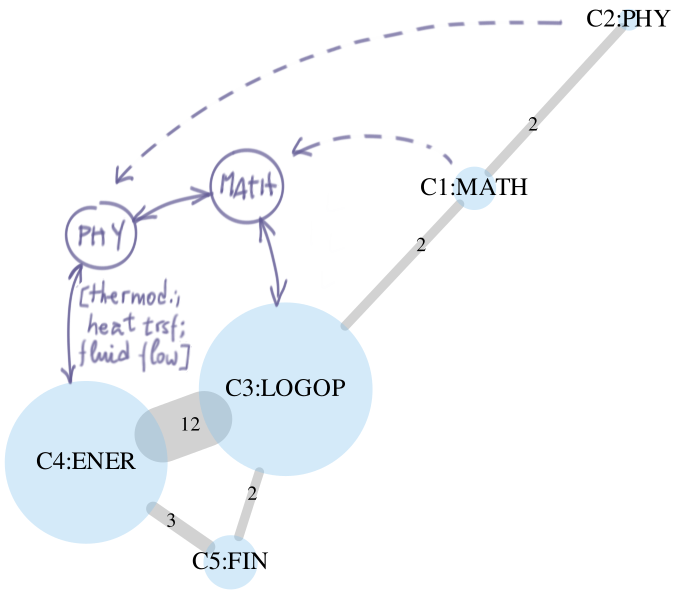}
  \captionof{figure}{Handwritten notes communicating a desirable change to the mapping in Figure \ref{fig:4.0} a). Example of how broader terms curriculum mapping can be used by stakeholders during the program planning discussions.}\label{fig:5.0}
\end{minipage}

\vspace{1\baselineskip}
Figure \ref{fig:5.0} demonstrates the ease with which stakeholders take possession of the mappings. The dashed lines show the preferred location of C2:PHY (and C1:MATH), closer to core courses. Text in square brackets points to broader terms justifying the link between C2:PHY and C4:ENER. Figure \ref{fig:5.0} could be the starting point for the revision of these courses LO; perhaps considering another forum and using detailed concept mappings.

To conclude this section, notice that having used a simple example with only 5 courses, it is not possible to verify if the method most frequent key program concepts and if the links between these and/ or courses are accurate. To assess the accuracy of the broader terms CM method, the next section presents a case study.

\section{The case study: Bachelor degree in T\&IM}\label{sec:4}
To evaluate the accuracy of the broader terms CM method, results obtained with this method should be confirmed by actual observations. For this purpose, this section uses a bachelor degree---Technology and Industrial Management (T\&IM)---assessed by the Portuguese accreditation agency (A3ES) in 2013. The section starts with the generic presentation of the T\&IM study program and with the presentation of the recommendations issued by A3ES (the results of the assessment). Afterwards, the broader terms CM method is used to generate mappings from courses LO-statements. To evaluate the accuracy of the method, the mappings are compared to the recommendations---which are deemed accurate. The section ends with a discussion of this comparison.

\subsection{T\&IM bachelor degree and the Portuguese accreditation agency recommendations}\label{sec:4.1}
The bachelor degree (180 ECTS credits) in Technology and Industrial Management (T\&IM, \citealp{Lourenco2013,Duarte2014,Duarte2018}) was conceived in 2006 at the College of Engineering of Instituto Politécnico de Setúbal, a Portuguese public HEI. The degree targeted mature students working in the industry sector in the region of Setúbal. Considering the characteristics of the students---mature blue color workers with formal and informal skills in their area of professional expertise---and the advanced technological settings provided by the employing organizations (which include automotive, aeronautic and ship repair industries), the 2007-2012 program plan emphasized managerial contents at the expense of engineering and mathematics. This emphasis on management topics is made clear in Figure \ref{fig:6.0}, a circular dendrogram representing T\&IM courses and respective departments. Out of the 38 program courses, 18 belonged to the Business Sciences Department.

\vspace{1\baselineskip}
\noindent%
\begin{minipage}{\linewidth}
  \center
  \includegraphics[scale=0.65]{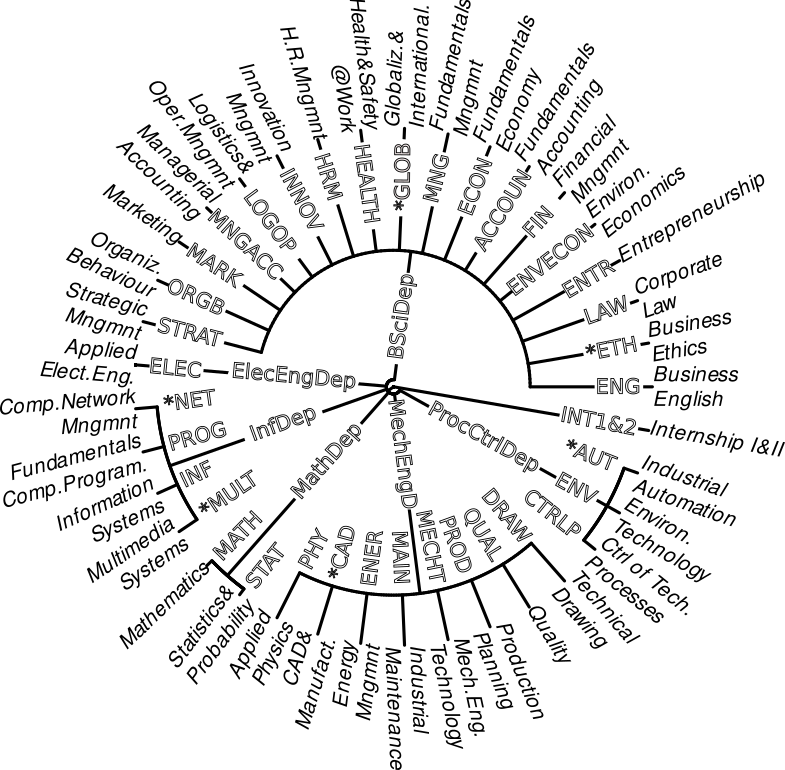}
  \captionof{figure}{Circular dendrogram representing T\&IM courses (2007-2012 program plan) and respective departments. Courses belonging to the departments of Business Sciences (BScDep), Electrical Engineering (ElecEngDep), Informatics (InfDep), Mathematics (MathDep), Mechanical Engineering (MechEngD) and Process Control (ProcCtrlDep) are represented counterclockwise. The responsibility for Internship I\&II is shared among departments and the asterisk symbol ($\ast$) is used to identify elective courses.}\label{fig:6.0}
\end{minipage}

\vspace{1\baselineskip}
Another important characteristic represented in Figure \ref{fig:6.0} is the dispersal of T\&IM core and elective courses among 6 departments.

Six years after it began, the Portuguese accreditation agency assessed the T\&IM bachelor degree \citep{A3ESurl2013}. Study program data reporting to the 2007-2012 period was gathered, a self-assessment report was delivered by the HEI, an independent panel of experts (representing A3ES) visited and met with HEI stakeholders.

Regarding the program plan, the A3ES produced the following recommendations:
\begin{enumerate}[i.]
\item Increase program-degree mathematical content.
\item Steer programming skills towards high-level languages with practical use.
\item Strengthen the program plan with important applied industrial management content, namely, in operations management, supply chain management and operational research.
\item Excessive number of courses, some with little additional content.
\item Poor integration of topics taught in the different courses.
\end{enumerate}

Concerning these recommendations, note that: (1) these are considered an \textit{accurate} expression of weaknesses in the 2007-2012 T\&IM program plan; (2) the non-prescriptive (and somewhat vague) style of the recommendations results from accreditation criteria allowing program-degrees to adjust to different HEI missions, to student demographics and available resources.

\subsection{T\&IM mappings}\label{sec:4.2}
Using LO-statements from the T\&IM courses (2007-2012 program) and the methodology described in Figure \ref{fig:1.0} (excluding the feedback loop), after courses LO classification and broader terms NLP, a total of 256 program tokens (no repetitions) was obtained. Figure \ref{fig:7.0} a) presents a wordcloud with the 200 most frequent key program concepts. Using the program biadjacency matrix $\mathbf{A}_\text{T\&IM}$, graphs with direct links between the most frequent key program concepts and with direct links between courses were obtained---Figures \ref{fig:7.0} b) and  \ref{fig:8.0}, respectively.

Note that out of the total 38 courses in Figure \ref{fig:6.0}, three (ETH, NET and CAD) were not taught and were excluded; Internships I\&II were also excluded, justifying the analysis of only 33 courses (for the meaning of the course acronyms please refer to Figure \ref{fig:6.0}).

\vspace{1\baselineskip}
\noindent%
\begin{minipage}{\linewidth}
  \center
  \includegraphics[scale=0.85]{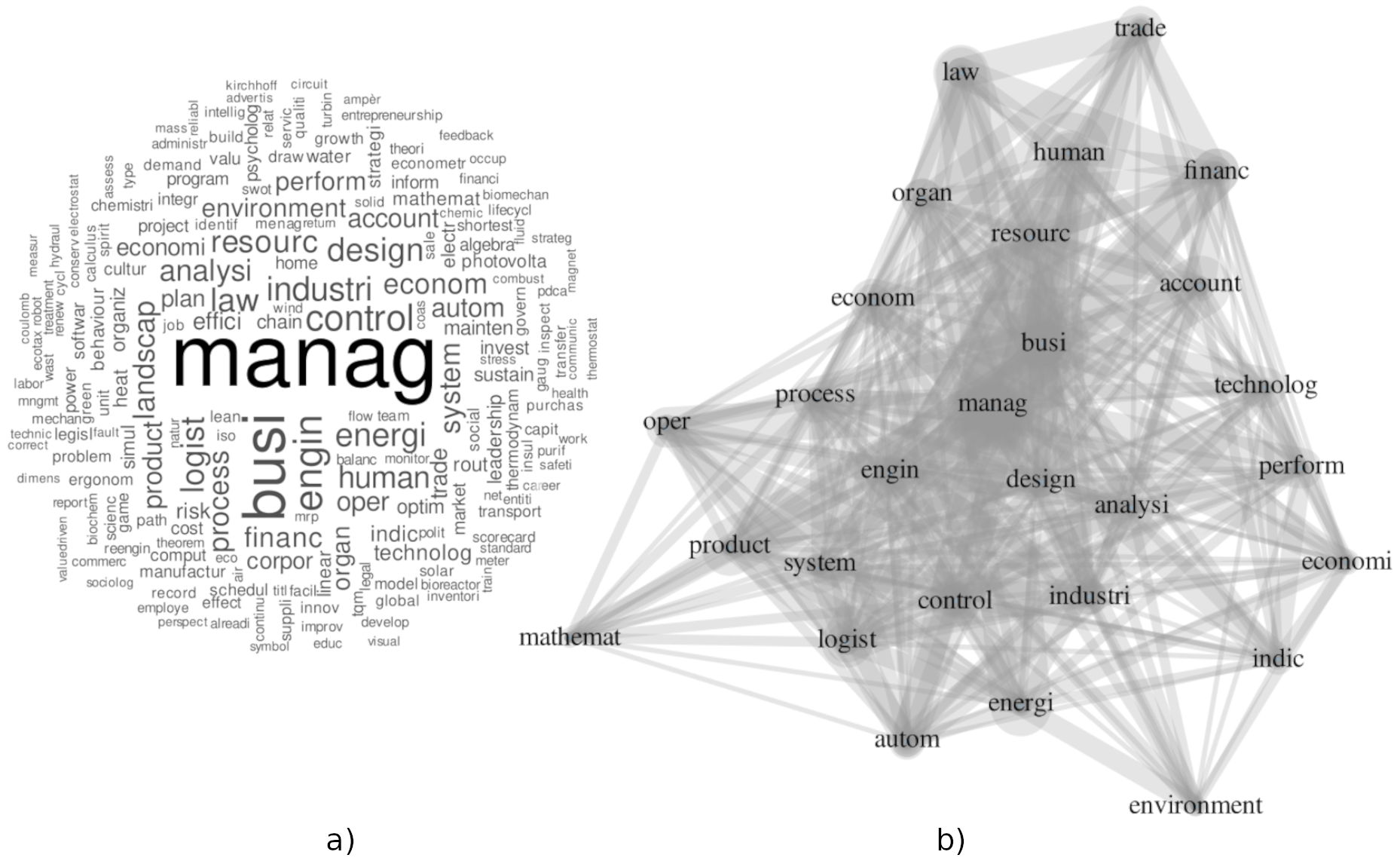}
  \captionof{figure}{Mappings for the T\&IM degree (2007-2012 program plan). a) Wordcloud with the 200 most frequent key program concepts. b) Links between the 28 most frequent key program concepts. Wordcloud obtained using the \texttt{wordcloud} package \citep{wordcloud2018}. Undirected network graph obtained from matrix \(\mathbf{L}_\text{28K}\) using the \texttt{iGraph} package \citep{igraph2006}. Both packages developed for the \texttt{R} programming language \citep{R2019}.}\label{fig:7.0}
\end{minipage}

\vspace{1\baselineskip}
With larger font size in Figure \ref{fig:7.0} a) and at the center of Figures \ref{fig:7.0} a) and \ref{fig:7.0} b) lay tokens ``manag[ement]'' and ``busi[ness]'', the most frequent key concepts found in the program broader terms. Besides ``manag[ement]'' and ``busi[ness]'', other key concepts lay in the vicinity of the graphs central region, namely, ``econom[y]'',   ``resourc[es]'', ``account'' (related to management); and, ``engi[neering]'', ``design'' (related to engineering). Because only 28 (out of 256) most frequent tokens are represented in Figure \ref{fig:7.0} b), all tokens exhibit a fair number of links. The way key concepts are linked in Figure \ref{fig:7.0} b) defines two distinct groups---or clusters---of key concepts: the management cluster, found towards the the top of the figure, and the engineering cluster at the bottom. Abstract key concepts such as ``process'', ``perform[ance]'', ``analysi[s]'' (``system'' or ``indic[es]'') are also found (mostly) in the interface between the management and engineering clusters.

\vspace{1\baselineskip}
\noindent%
\begin{minipage}{\linewidth}
  \center
  \includegraphics[scale=0.75]{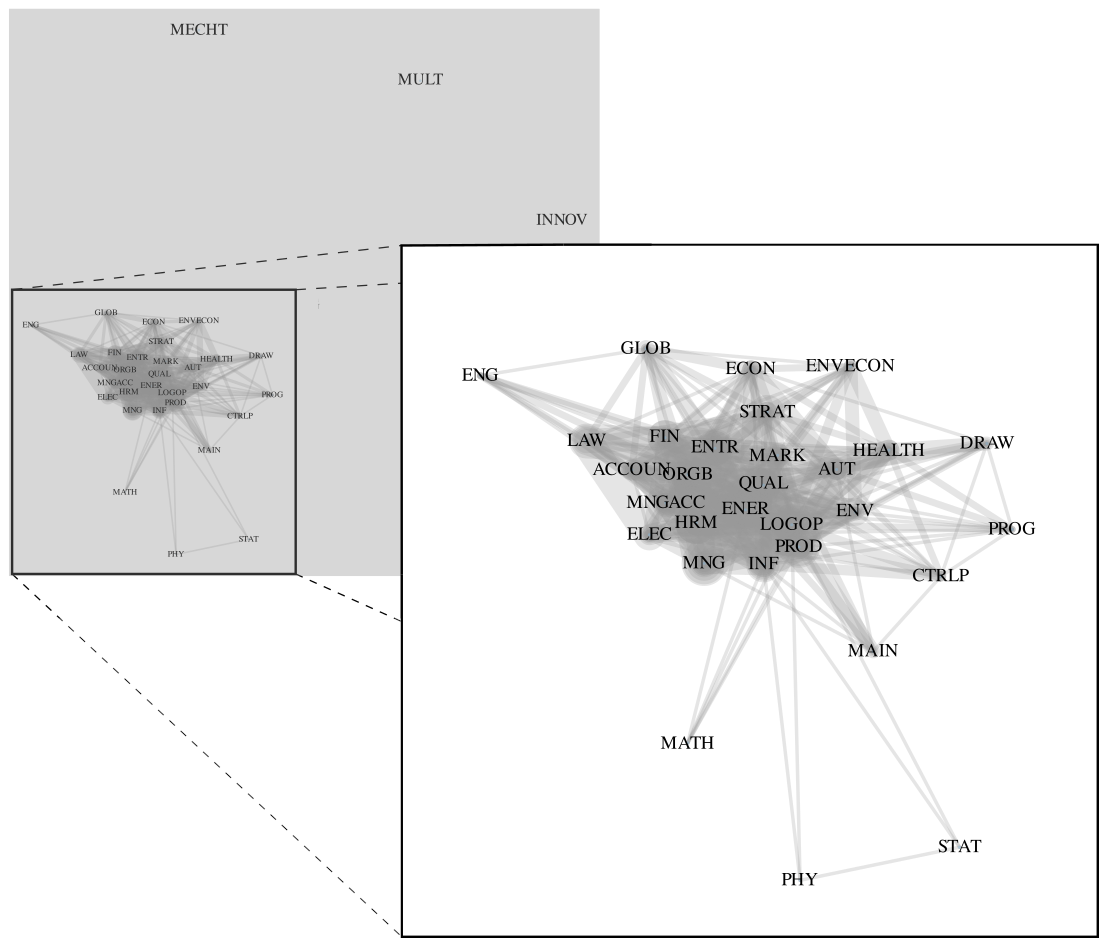}
  \captionof{figure}{Links between T\&IM program courses. The forward (white) plane presents an enlarged detail with 30 out of the 33 courses present in the backward (gray) plane. Network graph obtained from matrix \(\mathbf{L}_\text{33C}\) using the \texttt{iGraph} package \citep{igraph2006} for the \texttt{R} programming language \citep{R2019}. }\label{fig:8.0}
\end{minipage}

\vspace{1\baselineskip}
Figure \ref{fig:8.0} presents the links between program courses in two planes. The background (gray) plane is used to show three ultra-peripheral courses with no links: MECHT, MULT and INNOV. The forward (white) plane provides a detail of the courses laying closer to the graph core region. In this detail all courses are linked. Distant from the graph center lay courses PHY and STAT; at an intermediate distance lay MATH, MAIN, CTRLP, PROG, DRAW, ENVECON, ECON, GLOB and ENG; the remaining 19 courses lay at the central region. A divide similar to the one identified previously between managerial and engineering concepts is also present in Figure \ref{fig:8.0}, with managerial courses clustered to the (upper) left and engineering courses clustered towards the (lower) right of Figure \ref{fig:8.0} detail.

\subsection{Comparing T\&IM mappings with A3ES recommendations}\label{sec:4.3}
Comparing Figures \ref{fig:7.0} and \ref{fig:8.0} with A3ES recommendations (in Section \ref{sec:4.1}) it is possible to evaluate, for each recommendation, if the meaning conveyed in writing has a visual equivalent. The comparison of written and visual meaning is used to verify the accuracy of the broader terms CM method. 

Consider item (i) of the A3ES recommendations---increase program-degree mathematical content. The visual equivalent of this recommendation is the (relative) absence of mathematics-related tokens in the mappings. Indeed, Figure \ref{fig:7.0} a) includes very few mathematics-related tokens (e.g., mathemat, algebra, theorem), with the small font size of these tokens confirming the detachment of mathematics from core concepts taught in the T\&IM degree. The position of the ``mathemat'' token in Figure \ref{fig:7.0} b), distant from central key program concepts, is also consistent with this analysis. Figure \ref{fig:8.0} provides further evidence that some action should be taken concerning mathematics contents. Courses MATH and STAT relative position and the small number of links to other program courses translates into insufficient integration of mathematics contents.

As regards item (ii) of the A3ES recommendations---steer programming skills towards high level languages with practical use---, the visual equivalent should be the absence of links between programming and applied key concepts. An analysis similar to the previous one shows few programming-related tokens in Figure \ref{fig:7.0} a), with none among the 28 most frequent in Figure \ref{fig:7.0} b). As regards the PROG course, its location in Figure \ref{fig:8.0} confirms it is among those with less links to central and applied courses.

As for item (iii)---strengthen the program plan with important applied industrial management content, namely, in operations management, supply chain management and operational research---, from Figure \ref{fig:7.0} b), tokens ``oper[ational]'', ``logist[ics]'' are found among the 28 most frequent. Figure \ref{fig:7.0} a) includes additional concepts related to the mentioned courses, such as ``suppli'', ``chain'', ``optim''. Comparing font sizes in Figure \ref{fig:7.0} a), these latter key concepts are less frequent than generic managerial key concepts ``resourc'', ``financ'', ``account'', which could be subjectively deemed less important in a Technology \& Industrial Management program plan. A detailed quantitative analysis of token frequencies and of token connections could be made, contributing with relevant insights to the constructive discussion of this recommendation.

Using a similar line of inquiry, item (iv) in A3ES recommendations---excessive number of courses, some with little additional content---would benefit from the detailed analysis of token frequencies per course and from the equivalent to Figure \ref{fig:3.0} b) with data from the T\&IM study program. This detailed analysis and the graph are obtained with ease from matrix $\mathbf{A}_\text{T\&IM}$, using the methods and tool considered in supplementary material \citep{Duarte2020}. However, from Figure \ref{fig:8.0} it is possible to sort courses based on their connectivity (close to core or peripheral location). This figure depicts ultra-peripheral courses (MECHT, MULT, INNOV) in the background plane with no links. These courses are obvious candidates to detailed scrutiny. A scrutiny that should be extended to courses closer to the graph central region but, nevertheless, showing a small number of links (e.g., GLOB, ENVECON, ECON).

Finally, concerning item (v)---poor integration of topics---, as stated previously in Section \ref{sec:4.2}, Figures \ref{fig:7.0} b) and \ref{fig:8.0} denounce the clustering of managerial and of engineering concepts. In addition, courses more detached from the graph central region and with less links in Figure \ref{fig:8.0} (already identified in the previous A3ES recommendation, item iv) are once more obvious candidates to detailed scrutiny.

In light of the above, and considering A3ES recommendations, Table \ref{tab:2.0} (second column) summarizes the evidence-based visual meaning obtained from T\&IM mappings.

\vspace{1\baselineskip}
\noindent%
\begin{minipage}{\linewidth}
  \center
  \captionof{table}{Comparing A3ES recommendations with evidence-based visual meaning conveyed from T\&IM mappings.}\label{tab:2.0}
  \begin{tabular}{p{0.25cm}p{6.375cm}p{6.375cm}}
  \toprule
  \multicolumn{2}{l}{\textbf{A3ES recommendation}} & \textbf{Evidence from mappings}$^\mathbf{(a)}$ \\
  \midrule
  {\scriptsize i.} & {\scriptsize Increase program-degree mathematical content} & {\scriptsize Small number of mathematics-related key concepts and poor integration of mathematics-related courses} \\
  
  {\scriptsize ii.} & {\scriptsize Steer programming skills towards high-level languages with practical use} & {\scriptsize Extremely small number of programming-related key concepts and detached location of the programming course} \\

  {\scriptsize iii.} & {\scriptsize Strengthen the program plan with important applied industrial management content, namely, in operations management, supply chain management and operational research} & {\scriptsize Comparison of frequencies of applied industrial management key concepts with frequencies of generic managerial key concepts sheds light on the relative weight of each group in the program plan} \\

  {\scriptsize iv.} & {\scriptsize Excessive number of courses, some with little additional content} & {\scriptsize Identifies and sorts courses with few (and with no) links to core program courses.} \\

  {\scriptsize v.} & {\scriptsize Poor integration of topics taught in the different courses} & {\scriptsize Divide between engineering and management, visible both in key program concept and in course mappings} \\

  \bottomrule
  \multicolumn{3}{p{0.93\textwidth}}{\scriptsize (a) Note these results are obtained exclusively from courses LO-statements, whereas A3ES recommendations consider a visit by an independent panel of experts, interviews, focus group sessions, among other inputs.} \\
  \end{tabular}

\end{minipage}

\subsection{Discussion}\label{sec:4.4}
From Table \ref{tab:2.0}, for recommendations (i), (ii) and (v), mappings provide detailed visual evidence supporting these recommendations. For recommendations (iii) and (iv), the style (the vagueness) of A3ES statements prevents an objective comparison of visual and written meanings. Yet, these latter recommendations are useful to highlight the striking difference between an evidence-based analysis---possible with the mappings---and the subjective interpretation---relying on tacit understanding---of A3ES written statements.

Because the mappings provide evidence supporting the majority of the A3ES recommendations, it is concluded that the broader terms CM method provides an accurate depiction of T\&IM program plan weaknesses. Because all T\&IM mappings rely on key program concepts, it is also concluded that these key concepts---and the broader terms CM method---are useful in program planning.

Three additional notes are worth mentioning. Firstly, despite the large number of program courses (33), classification, NLP and visualization steps were concluded quickly and with ease, posing no particular difficulty. Secondly, Figure \ref{fig:8.0} shows that a holistic experience of the T\&IM program plan, considering interrelations between the 33 courses, is possible. Lastly, unlike the \textit{course} mapping of \citet{Meij2018}, the detailed \textit{concept} mappings of \citet{Seering2015}, \citet{Willcox2017} or \citet{Varagnolo2020}, visual outputs from the broader terms CM method do not aim at the \textit{tracing} of the available learning pathways or at the \textit{tracing} of detailed precedence relations between program concepts. Instead, the broader terms CM method \textit{maps}\footnote{\citet{Wang2015}, based on views derived from Gilles Deleuze and Félix Guatarri, discusses the distinction between mapping and tracing in the context of curriculum mapping. This researcher supports that current practice of curriculum mapping in higher education is, actually, tracing. Current curriculum mappings represent fixed routes with a linear tree-like structure and an objective model of the curriculum, and this is an example of tracing. Maps have different topological characteristics. Like rhizomes, maps do not aim at guiding to a main road or familiar destination, but to represent the mesh of nodes and the patterns that emerge through the multitude of connections between nodes.} clusters of key concepts, or courses; maps multiple undirected links between courses and/ or concepts. These maps' aim is to provide a representation of the program plan that is understandable to all stakeholders, allowing participatory involvement or non-faculty groups without imposing predefined models or fixed routes. In this sense, broader terms curriculum mapping does not replace, rather precedes and complements other curriculum mapping methods,

\section{Conclusion}\label{sec:5}
Addressing the curriculum development process is of paramount importance. This process has profound consequences being responsible for the preparation of future professionals and for laying the foundations for dynamic knowledge transfer systems affecting local and global realities. At the heart of curriculum development lays program planning. Program planning is of immense strategic value. The effort put into program planning propagates through all levels and subprocesses of teaching and learning, imprinting the values, intentions and expectations that will guide stakeholders; shaping HEI educational outcomes.

To improve program planning more participatory touchpoints to non-faculty groups (i.e., students, industry, society) are needed. Creating these touchpoints, contributing to representative program planning was the motivation behind this paper.

An important impediment to representative program planning lays in the communication gap between faculty and non-faculty groups. Curriculum mapping has been used to promote better communication between faculty and shape program planning. This paper collected practices available from different types of curriculum mapping and, using information and data science techniques, tailored a curriculum mapping method for non-faculty groups participation in program planning discussions. The resulting method---the broader terms CM (Curriculum Mapping) method---was illustrated with the help of a simple example---5 courses example. The following conclusions were found:
\begin{itemize}
\item (Section \ref{sec:3.1}) Classification replaces head-on discussion of subjective course LO-statements with the much more objective task of selecting broader terms from a control vocabulary.
\item (Section \ref{sec:3.2}) Natural language processing allows the  quantitative analysis of the program plan, providing a way to cut across disciplinary and scientific boundaries, hierarchical and functional differences, and atomized discourses.
\item (Section \ref{sec:3.3}) Mappings render quantitative results' interpretation universal, self-explanatory, empowering stakeholders with evidence-based details on weaknesses (and strengths) in the program plan.
\item (Section \ref{sec:3.4}) Discussion of visual outputs with non-faculty groups allows representative program planning, with these groups' voice being heard on reclassification and review of course LO-statements.
\end{itemize}

Despite the relevance of the above conclusions---related to the participatory involvement of non-faculty stakeholders---, the simple 5 courses example was unable to answer the question of the broader terms CM method' accuracy and, therefore, of the method' utility.

To evaluate the method's accuracy, a case study---the T\&IM bachelor degree---was used. Mappings for the case study were obtained and compared with observations from an independent panel of experts. From this comparison the following was concluded (Section \ref{sec:4.4}):
\begin{itemize}
\item Mappings provide evidence supporting the observations, and the broader terms CM method provides an accurate depiction of T\&IM program plan weaknesses.
\item Key concepts obtained from course LO-statements---and the broader terms CM method---are useful in program planning.
\end{itemize}

Considering the benefit of non-faculty groups' participation in curriculum development processes, and considering the progresses made in information systems and relational databases \citep{Chen1976,Bagui2003,Leff2001}, the merger of techniques used in the broader terms CM method and HEI information systems would help bring the method's benefits into HEI everyday reality; for example, with the inclusion of mappings in information systems' summary dashboards. This merger is just one potential topic for further explorations in this rich and challenging research area that joins education, information and data sciences.

\vfill

\pagebreak
\bibliographystyle{unsrtnat}

\end{document}